\documentclass{article}

\usepackage[preprint,nonatbib]{neurips_2026}
\usepackage{pslatex}
\usepackage{hyperref}
\usepackage{lineno}
\usepackage{xcolor}
\usepackage{bookmark}
\usepackage{graphicx}
\usepackage{tikz}
\usepackage{amsmath}
\usepackage{amssymb}
\usepackage{amsthm}
\usepackage{mathtools}
\usepackage{placeins}
\usepackage{bbm}
\usepackage[capitalize,noabbrev]{cleveref}
\usepackage{enumitem}
\usepackage{booktabs}
\usepackage{wrapfig}
\usepackage{caption}
\usepackage{soul}
\usepackage{multirow}
\usepackage{makecell}
\usepackage{array}
\usepackage{subcaption}
\usepackage{todonotes}
\usepackage{changes}
\usepackage{setspace}
\usepackage{float}
\usepackage{floatpag}
\usepackage{xltabular}
\hypersetup{colorlinks=true, linkcolor=red!90!black, citecolor=green!50!black, urlcolor=blue!90!black}
\usepackage[style=authoryear,backend=biber,natbib,giveninits=true,uniquename=init,sorting=nyt,maxcitenames=1,mincitenames=1,minbibnames=6,maxbibnames=6,hyperref=true]{biblatex}
\DeclareFieldFormat{bibhyperref}{#1}
\DeclareFieldFormat{fullcitehyperref}{\bibhyperref{#1}}
\newbibmacro*{cite:full}{%
    \iffieldundef{shorthand}
    {\ifboolexpr{test {\ifnameundef{labelname}} or test {\iffieldundef{labelyear}}}
        {\usebibmacro{cite:label}\setunit{\printdelim{nonameyeardelim}}}
        {\printnames{labelname}\setunit{\printdelim{nameyeardelim}}}%
    \usebibmacro{cite:labeldate+extradate}}
{\usebibmacro{cite:shorthand}}}
\newbibmacro*{citeyear:bare}{%
    \iffieldundef{shorthand}
    {\iffieldundef{labelyear}
        {\usebibmacro{cite:label}}
    {\usebibmacro{cite:labeldate+extradate}}}
{\usebibmacro{cite:shorthand}}}
\renewbibmacro*{citeyear}{%
\printtext[fullcitehyperref]{\usebibmacro{citeyear:bare}}}
\renewbibmacro*{cite}{%
\printtext[fullcitehyperref]{\usebibmacro{cite:full}}}
\renewbibmacro*{textcite}{%
    \printtext[fullcitehyperref]{%
        \ifnameundef{labelname}
        {\iffieldundef{shorthand}
            {\usebibmacro{cite:label}%
                \setunit{\global\booltrue{cbx:parens}\printdelim{nonameyeardelim}\bibopenparen}%
                \ifnumequal{\value{citecount}}{1}{\usebibmacro{prenote}}{}%
            \usebibmacro{cite:labeldate+extradate}}
        {\usebibmacro{cite:shorthand}}}
        {\printnames{labelname}%
            \setunit{\global\booltrue{cbx:parens}\printdelim{nameyeardelim}\bibopenparen}%
            \ifnumequal{\value{citecount}}{1}{\usebibmacro{prenote}}{}%
\usebibmacro{citeyear:bare}}}}

\definecolor{sigpurple}{HTML}{674EA7}
\definecolor{siggreen}{HTML}{38761D}
\newcommand{\purp}[1]{\textcolor{sigpurple}{#1}}
\newcommand{\grn}[1]{\textcolor{siggreen}{#1}}
\newcommand{\hfmodel}[1]{\href{https://huggingface.co/#1}{\nolinkurl{#1}}}
\newcommand{\hfmodelshort}[2]{\href{https://huggingface.co/#1}{\nolinkurl{#2}}}

\title{
    What Makes Two Language Models Think Alike?
}

\author{%
    Louis~Jalouzot \\
    UNICOG, CNRS, INSERM, CEA, Paris-Saclay University\\
    LSCP, EHESS, ENS, CNRS, PSL University\\
    \texttt{jalouzot.louis@gmail.com} \\
    \And
    Christophe~Pallier \\
    UNICOG, CNRS, INSERM, CEA, Paris-Saclay University\\
    \And
    Emmanuel~Chemla \\
    LSCP, EHESS, ENS, CNRS, PSL University\\
    Earth Species Project\\
    \And
    Yair~Lakretz \\
    LSCP, EHESS, ENS, CNRS, PSL University\\
}

\begin{document}



\maketitle



\begin{abstract}
    Do architectural and training differences influence the way models represent and process language?
    Traditional similarity metrics tell us whether two models share a similar representational geometry, but they cannot explain \textit{why}.
    Here, we propose a new, simple, approach to address this question. This approach maps neural activity in each model layer onto a set of interpretable linguistic features and quantifies how much each of them drives similarities and differences between models.
    We use this approach to compare 43 language models across 10 families, including decoder Transformers, State-Space Models, and Recurrent Neural Networks.
    We find that model-level similarity is driven most strongly by release date, a proxy for general LLM development, and model family, suggesting that linguistic signatures are not primarily shaped by scale or architecture class.
    Overall, our approach provides a way to link theoretically-motivated symbolic descriptions to neural representations and can readily be extended to other domains such as speech and vision, and to other neural systems such as biological brains.


\end{abstract}

\begin{quote}
    \small
    \textbf{Keywords:} Neural Representations; Metric Learning; Large Language Models; Interpretability; Symbolic and non-Symbolic approaches
\end{quote}


\section{Introduction}
\label{sec:introduction}

Marr's hierarchy describes information-processing systems using three levels (\cref{fig:marr}; \citealp{marr_vision_2010}): computational, algorithmic, and implementational.
The computational level defines the system's goal.
Many algorithms can solve the same computational goal.
Finally, the implementational level concerns how an algorithm is realized.
In biological systems, this includes physical hardware such as brains.
In artificial systems, it also includes network topology and architecture.
Many implementations can run the same algorithm.

Large Language Models (LLMs) fit into Marr's hierarchy.
At the computational level, most language models share the same goal: next-token prediction (but see \citealp{gloeckle_better_2024} for multi-token prediction).
At the implementational level, models can use very different architectures, such as Transformers, State-Space Models (SSMs), or Recurrent Neural Networks (RNNs).
At the algorithmic level, the Platonic Representation Hypothesis \citep{huh_platonic_2024} predicts increasing convergence: as models become larger and more competent, their representations should align across architectures, objectives, and even data modalities.
From Marr's perspective, different implementational choices can still lead to shared algorithmic organization.
Yet differences in architecture and training recipes may also make models rely on distinct linguistic representations.
Empirical work on this hypothesis has primarily used geometric alignment measures, such as mutual nearest-neighbor overlap between representational kernels, leaving open whether convergent representations are organized by the same explicit linguistic features.
In this work, we ask whether language models converge at this algorithmic level, and if so which linguistic features explain that convergence.

\setlength{\columnsep}{0.5cm}
\begin{wrapfigure}{r}{0.5\linewidth}
    \centering
    \includegraphics[page=4, trim=0 1.5cm 0 0, width=\linewidth]{figures/mlem_methods.pdf}
    \caption{
        \textbf{Marr's levels of analysis:}
        While many language models share the computational goal of next-token prediction (top level), their architectures and training might differ substantially (bottom level: Transformer, SSM, RNN).
        Given the one-to-many relationship between each level of Marr's hierarchy to the level below it, it is unclear whether distinct LLM architectures (bottom level) would develop the same representations and algorithms (middle level) to perform the computational task (top level).
    }
    \label{fig:marr}
    \vspace{-0.25cm}
\end{wrapfigure}
\setlength{\columnsep}{0.5cm}

Standard model-similarity measures compare raw representational geometry, including second-order isomorphism/RSA \citep{shepard_secondorder_1970,laakso_content_2000,mehrer_individual_2020,kriegeskorte_representational_2008,abnar_blackbox_2019}, linear regression \citep{romero_fitnets_2015}, CCA \citep{raghu_svcca_2017,morcos_insights_2018,wu_similarity_2020,belinkov_analysis_2019}, statistical shape analysis \citep{williams_generalized_2022}, and DSA \citep{ostrow_geometry_2023}.
These measures tell us \textit{whether} two models share a similar raw geometric space.
However, they act as a black box.
They do not explain \textit{why} the representations align.
They cannot tell us if two models are similar because, e.g., they both heavily encode the same syntactic structure, or simply because they share superficial lexical biases.

To address this, we shift from purely geometric comparisons to feature-based representational similarity.
We use metric-learning encoding models (MLEMs; \cite{jalouzot_metric_2025}).
MLEMs learn a metric over interpretable features of stimuli to approximate the distances between the neural representations of those stimuli.
Applied to embeddings of language models on sentences with specific linguistic properties, this produces a \textit{linguistic signature}, a profile that quantifies how much each linguistic feature contributes to predicting neural distances among stimuli for a specific layer or model.
By comparing these signatures, we measure similarity based on explicit linguistic strategies, rather than raw geometry.

We test this approach on 43 models across three different classes of neural architectures: decoder Transformers, State-Space Models (SSMs), and Recurrent Neural Networks (RNNs).
We find that the similarity between model linguistic signatures is driven most strongly by their family and release date.
The latter acts as a proxy for general LLM development, summarizing correlated changes in architecture and training recipes.
Parameter count, training token count, and depth make smaller contributions.
This suggests that the way language models solve language processing is shaped more by model family and model generation than by scale or specific architectural differences.

More broadly, the same feature-based comparison can extend to speech, vision, or brain recordings whenever stimuli can be paired with explicit theoretical features.


\section{Methodology: Linguistic Signature Similarity}
\label{sec:methodology}

\subsection{Measuring Similarity through Interpretable Features}
Two standard approaches relate theoretical features to neural representations: decoding and encoding models \citep{kriegeskorte_interpreting_2019}.
Decoding models, or probes, predict features from neural activity \citep{hupkes_visualisation_2018,tenney_what_2019,arps_probing_2022}.
High decoding performance shows that feature-related information is available in the representations, but it does not show that the feature itself organizes the neural space.
A feature can be decodable simply because it correlates with another variable actually used by the model \citep{hewitt_designing_2019,belinkov_probing_2022,kumar_probing_2022}.
Encoding models instead predict neural activity from features and can include several features or confounds at once, disentangling their contributions.
However, standard encoding models are often univariate: they fit each neuron, voxel, or embedding dimension independently.
They can therefore miss information encoded only in distributed geometry across a population of units \citep{georgopoulos_neuronal_1986,rumelhart_parallel_1986}.
MLEMs keep the encoding logic but move the target from individual activations to pairwise neural distances among stimuli.
They learn which feature distances best predict those neural distances, yielding feature importances for the geometry of a layer's representations.
In our approach, these importances define the layer's \textit{linguistic signature}.
We then compare linguistic signatures across layers and models.
If two models or layers have similar signatures, their representations are organized by similar linguistic variables, even when their raw geometries differ.

\begin{figure*}[p!]
    \vspace*{\fill}
    \centering
    \begin{subfigure}{\textwidth}
        \centering
        \includegraphics[page=5, trim=0 4cm 0 0, width=\linewidth]{figures/mlem_methods.pdf}
        \caption{General MLEM Pipeline (figure adapted from \citealp{jalouzot_metric_2025})}
        \label{fig:mlem}
    \end{subfigure}
    \\[0.5cm]
    \begin{subfigure}{.5\textwidth}
        \centering
        \includegraphics[page=7, width=.9\linewidth]{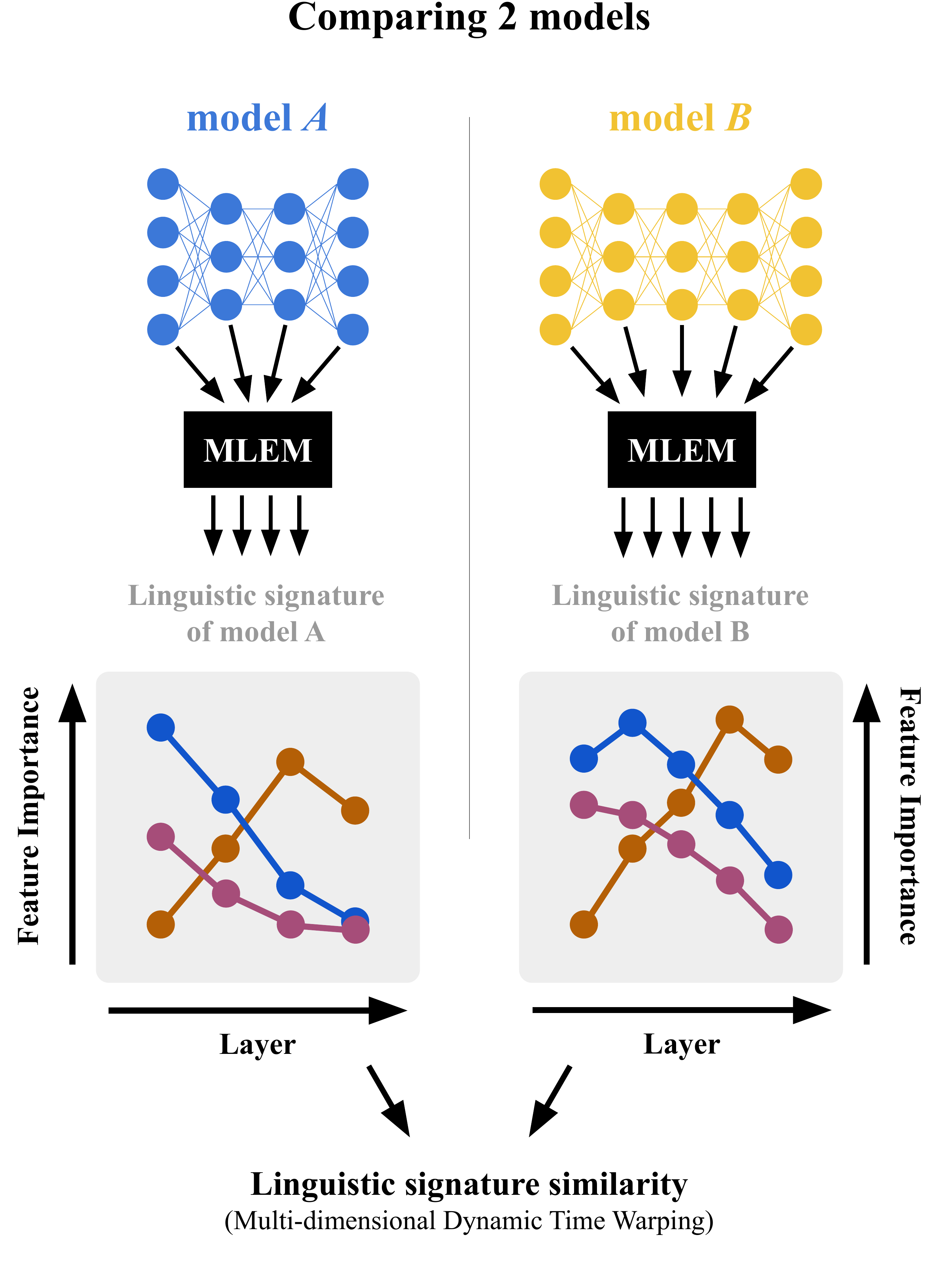}
        \caption{Model Comparison}
        \label{fig:model_comparison}
        \vfill
    \end{subfigure}
    \hfill
    \begin{subfigure}{.45\textwidth}
        \centering
        \includegraphics[page=6, width=.9\linewidth]{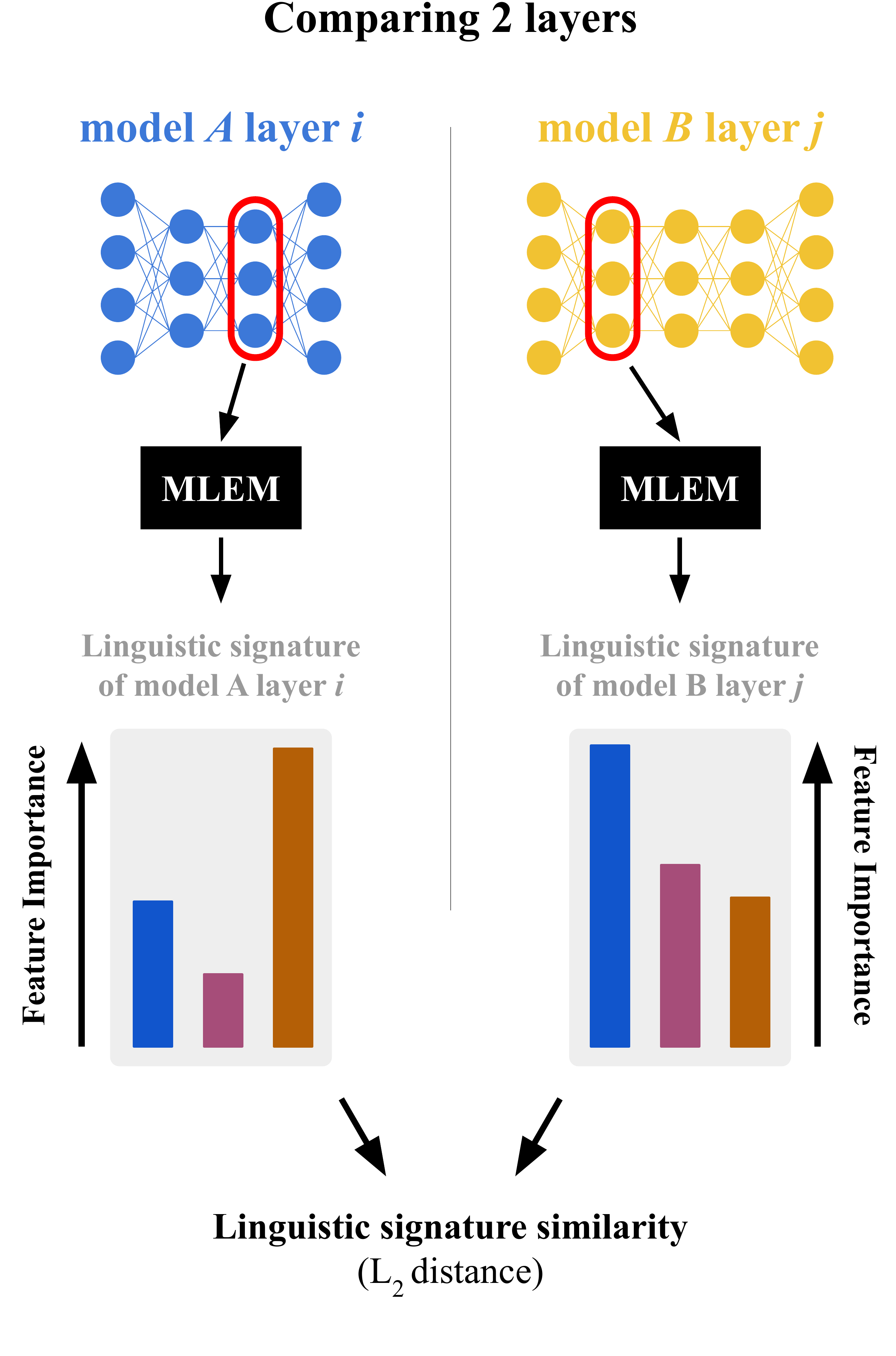}
        \caption{Layer Comparison}
        \label{fig:layer_comparison}
        \vfill
    \end{subfigure}
    \\[0.5cm]
    \caption{
        \textbf{Methodological Approach}:
        (a) General MLEM Pipeline: MLEMs identify which interpretable stimulus features organize neural representations by testing which \purp{feature distances} predict \grn{neural distances} among stimuli.
        To do so, they start from stimuli paired with features and neural representations, predict \grn{pairwise neural distances $D^N$} from \purp{pairwise feature-based distances $D^f$}, and compute the importance of each feature.
        We apply MLEM to a synthetic dataset of sentence stimuli with corresponding linguistic features and neural representations from the layers of language models.
        (b) Model Similarity: To compare whole language models, we ask whether they assign similar importance to linguistic features across layers.
        We represent each model by the layer-wise profile of its linguistic feature importances and compare these profiles with multi-dimensional Dynamic Time Warping.
        (c) Layer-Wise Similarity: To compare individual layers, we ask whether two layers assign similar importance to the same linguistic features.
        We define each layer's \textit{linguistic signature} as its vector of linguistic feature importances and compare these vectors with Euclidean distance ($L_2$).\\
        This approach yields an interpretable similarity measure based on linguistic priorities rather than raw geometry alone.
    }
    \label{fig:pipeline}
    \vspace*{\fill}
\end{figure*}

\subsection{Metric-Learning Encoding Models (MLEMs)}
\label{sec:mlem}

MLEMs \citep{jalouzot_metric_2025} learn a distance function over input features (e.g., tense, subject number) to match the actual distances between neural representations (\cref{fig:pipeline}).

Formally, consider a set of $n$ stimuli (e.g. sentences), each described by $m$ interpretable features $F$ (e.g., tense, gender).
First, we consider the \textit{pairwise neural distances} $D^N$ between the neural embeddings $Y_i$ and $Y_j$ corresponding to stimuli $s_i$ and $s_j$ (for instance with the Euclidean distance)\footnote{
    Many neural dissimilarities could be used such as
    ones based on cosine similarity, correlation, or Mahalanobis distance depending on the representation type and noise structure \citep{kriegeskorte_representational_2008,diedrichsen_representational_2017}.
    We choose the classical Euclidean distance for its simplicity and interpretability.
}: $D_{ij}^N = \left\|Y_i - Y_j\right\|_2$.
Second, we consider the \textit{pairwise feature distances}.
Let $D_{ij}^F$ be the $m$-dimensional vector of feature distances for a pair of stimuli: $D_{ij}^F=\left( \ D_{ij}^{f_1} \ , \ D_{ij}^{f_2} \ , \ \ldots \ , \ D_{ij}^{f_m} \ \right)$, where $D_{ij}^{f_k}$ is the distance with respect to feature $f_k$.
For example, for a categorical feature such as sentence tense, $D_{ij}^{f_k}=0$ if two sentences have the same tense and $D_{ij}^{f_k}=1$ otherwise.
For an ordered feature such as word frequency, $D_{ij}^{f_k}$ can instead be the absolute difference between the two feature values.
To evaluate which input features best explain the neural distances, MLEMs learn a distance over the feature-distance vectors $D_{ij}^F$ that best models the neural distances $D_{ij}^N$.
Formally, they learn a symmetric positive definite matrix $W \in \mathbb{S}^{++}_m$ such that $D_{ij}^N$ is modeled as $\widehat{D}_{ij}^N=\left\|D_{ij}^F\right\|_W = \sqrt{{D_{ij}^F}^T W D_{ij}^F}$.

The diagonal entries of $W$ weigh individual features, while its off-diagonal entries weigh feature interactions.
MLEMs optimize $W$ to maximize the Spearman correlation $\rho$ between the empirical neural distances $D_{ij}^N$ and the modeled distances $\widehat{D}_{ij}^N$: $\sup_{W\in\mathbb{S}^{++}_m} \rho_{i<j}\left( \ \widehat{D}_{ij}^N \ , \ D_{ij}^N \ \right)$.

The Spearman correlation focuses on preserving the relative arrangement (topology) of stimuli, rather than the absolute scale which is arbitrary in neural spaces.
Because exact ranks are not differentiable, the model optimizes a differentiable relaxation of $\rho$ \citep{blondel_fast_2020}.
The model is then trained using stochastic gradient descent on batches of stimulus pairs.
Positive definiteness of $W$ is enforced via Cholesky parameterization.

\subsection{Extracting Linguistic Signatures (Feature Importance Profiles)}
Feature Importances (FI) for each interpretable feature can be computed from an MLEM.
Feature Importance is defined as the average decrease in the model's Spearman correlation score on a test set when randomly permuting a specific feature's distance matrix or that of an interaction between two features.
This approach is more robust than looking at the raw weights in $W$, which can be more affected by correlations between features and scaling \citep{breiman_random_2001}.
The scale of FIs is more interpretable as they are differences of Spearman scores and in particular it allows for comparison between different MLEMs (e.g. trained on neural representations from different language models).

For each layer of a language model, we collect sentence-level embeddings from the last-token representation of each sentence in a synthetic dataset balanced to mitigate correlations among linguistic features of interest (described later), and compute the FIs for those features using MLEMs.
We refer to the resulting FI vector as the layer's \textit{linguistic signature}.
A high FI means that differences along that feature predict relatively large distances between sentence embeddings in that layer.
For example, if tense has high FI, sentence pairs that differ in tense tend to be farther apart in the layer's representation space than sentence pairs with the same tense.
For an entire model, we consider the layer-wise profile formed by the linguistic signatures of its different layers and refer to this profile as the model's \textit{linguistic signature}.
It tracks the development of linguistic features along its layers.

The FIs are computed on held-out sentences using 5-fold cross-validation.
Throughout the different analyses, we report the average of the metric of interest across those folds and display the variability (standard deviation) as error bars or shaded areas to help reliably compare data points.

\subsection{Linguistic Signature Similarity} \label{sec:linguistic_signature_similarity}
Because a linguistic signature summarizes how strongly a layer relies on each linguistic feature, comparing signatures lets us move beyond standard geometric similarity.
The question is not only whether two representations are close, but whether they prioritize the same linguistic information.
We analyze linguistic-signature similarity at three levels: model-level, layer-level, and feature-level.

At the \textbf{model level}, we want to compare how linguistic signatures evolve across layers, but models often have different numbers of layers.
A direct layer-by-layer comparison is therefore not well defined, and even models with the same number of layers may express similar feature-importance patterns at shifted relative layer positions.
To address this, for each pair of models we compute normalized multi-dimensional \textbf{Dynamic Time Warping} (DTW; \cite{sakoe_dynamic_1978}) between their layer-wise FI profiles (\cref{fig:model_comparison}).
DTW aligns similar feature-importance vectors along the depth axis, allowing us to test whether linguistic features follow similar profiles across layers even when models differ in layer count or features emerge at different depths.
The normalization further mitigates the confounding effect of large differences in numbers of layers between models.
Implementation details can be found above \cref{fig:dtw_heatmap} in Appendix.

We then fit a second MLEM on the resulting model-level similarity matrix to explain its structure from model properties.
The predictors are family membership, architecture class (Transformer, SSM, RNN), parameter count, release date, depth, depth-to-width ratio\footnote{We call \textit{depth} the number of layers and \textit{width} the number of hidden units per layer.}, and training token count.
Numerical properties that span multiple orders of magnitude (parameter count, depth, and training token count) are log-scaled
We included release date as a proxy for general LLM development.
It summarizes many architectural choices (e.g., activation type, normalization type, positional encoding, attention type) which are highly correlated with one another, as well as factors we cannot easily retrieve or estimate (e.g., training FLOPs, training data types).
We also excluded width and vocabulary size because they were highly correlated with the other variables.
The remaining model properties show only moderate pairwise correlations (\cref{fig:dtw_feature_corrs} in Appendix), making the resulting FI profile more reliable.

At the \textbf{layer level}, we compare individual FI profiles using Euclidean distance (\cref{fig:layer_comparison}).
Two layers are close only if they rely on the same linguistic features with the same intensity.

At the \textbf{feature level}, we inspect why particular layer pairs are close or far.
We compare their linguistic signatures in scatter plots to find which features lead the similarity or difference.
We visualize the neural representations at those layers using Multi-Dimensional Scaling (MDS; \cite{kruskal_multidimensional_1964}) to study whether in fact the models organize their neural space with respect to the leading features in a similar way or not.
This provides a qualitative validation that linguistic-signature similarity reflects actual neural representations similarity with respect to the identified linguistic features.

Finally, we compare this feature-based approach at the layer level to a feature-agnostic baseline.
We use Representational Similarity Analysis (\textbf{RSA}; \citealp{kriegeskorte_representational_2008}) directly on the pairwise neural distance matrices $D^N$ for each layer.
Contrary to RSA, linguistic-signature similarity is more interpretable and naturally supports the three complementary levels of analysis above.

\section{Experimental Setup}
\label{sec:experimental_setup}

We evaluated 43 autoregressive language models from 10 families, spanning 100M to 14B parameters and 3 architecture classes: decoder Transformers, State-Space Models, and Recurrent Neural Networks.
The model set includes older Transformer baselines (GPT-2, OPT, Pythia), newer Transformer families (OLMo-2, Llama-3.2, Ministral-3, Qwen3), State-Space Models (Mamba, Mamba-2), and a Recurrent Neural Network family (RWKV-7).
All models share the next-token prediction objective but differ in architecture, scale, training data, and release date, allowing us to test which factors explain linguistic-signature similarity.
For each model and layer, we collected the hidden state of the last token of each sentence, as it is the position that has access to the full sentence in autoregressive models.

We used the Relative Clause dataset from \cite{jalouzot_metric_2025} which contains 7.7k sentences generated using Context-Free Grammars.
The dataset crosses relative clause attachment site (center-embedded vs. peripheral) with relative clause type (subject vs. object), and provides 12 linguistic features including syntactic properties, and lexical frequency and morphological attributes of particular elements of the sentence.
Its balanced design mitigates feature correlations (\cref{fig:rc_correlations}), which is important for reliable Feature Importances \citep{breiman_random_2001}.

More details on the included models, embedding extraction procedure, and dataset design are provided in \cref{sec:model_inventory,sec:relative_clause_dataset} in Appendix.


\begin{figure*}[ht!]
    \centering
    \begin{subfigure}[c]{.65\linewidth}
        \centering
        \includegraphics[width=\linewidth]{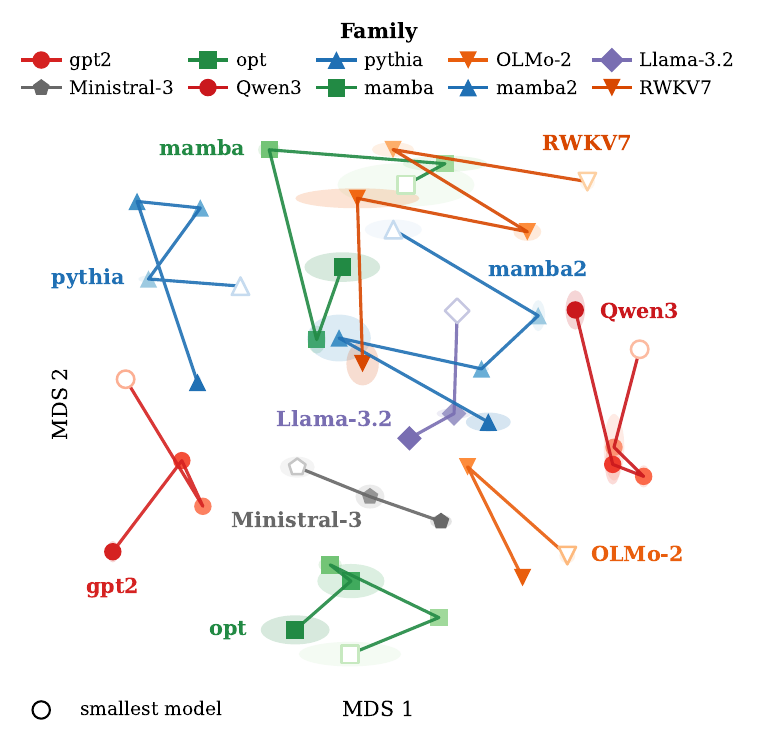}
        \caption{MDS projection of similarity between model linguistic signatures.}
    \end{subfigure}
    \hfill
    \raisebox{-0.5\height}{\rule{0.1pt}{6cm}}
    \hfill
    \begin{subfigure}[c]{.3\linewidth}
        \centering
        \includegraphics[width=\linewidth]{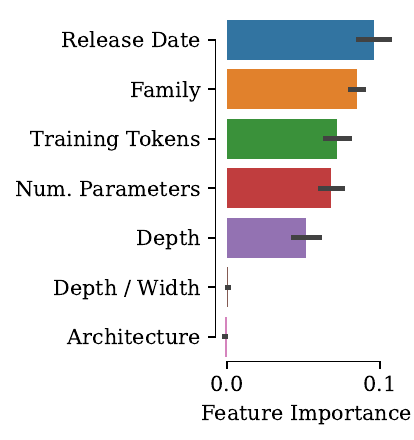}
        \caption{
            Feature importance of model properties for explaining linguistic signature similarities.
            Error bars show standard deviation across the original 5-fold cross-validation used to estimate layer-wise linguistic signatures.
        }
        \label{fig:dtw_fi}
    \end{subfigure}
    \caption{
        \textbf{What Makes Two Language Models Think Alike?}
        (a): MDS visualization of model linguistic signature similarities computed using Dynamic Time Warping.
        Each marker corresponds to an LLM.
        Successive LLMs from the same family ordered by size are connected by a line.
        Models from the same family group together, highlighting that they share similar linguistic feature importance profiles across layers.
        The DTW distance matrix is shown in \cref{fig:dtw_heatmap} in Appendix.
        (b): To identify which model properties best predict distances among model linguistic signatures, we fit an MLEM to the DTW distance matrix and compute Feature Importances for those properties.
        This analysis shows that model-level distances are explained mainly by family membership and release date, highlighting the role of correlated architectural changes from different LLM generations.
    }
    \label{fig:dtw_similarity}
    \vspace{-0.25cm}
\end{figure*}

\begin{figure*}[ht!]
    \centering
    \includegraphics[width=\linewidth]{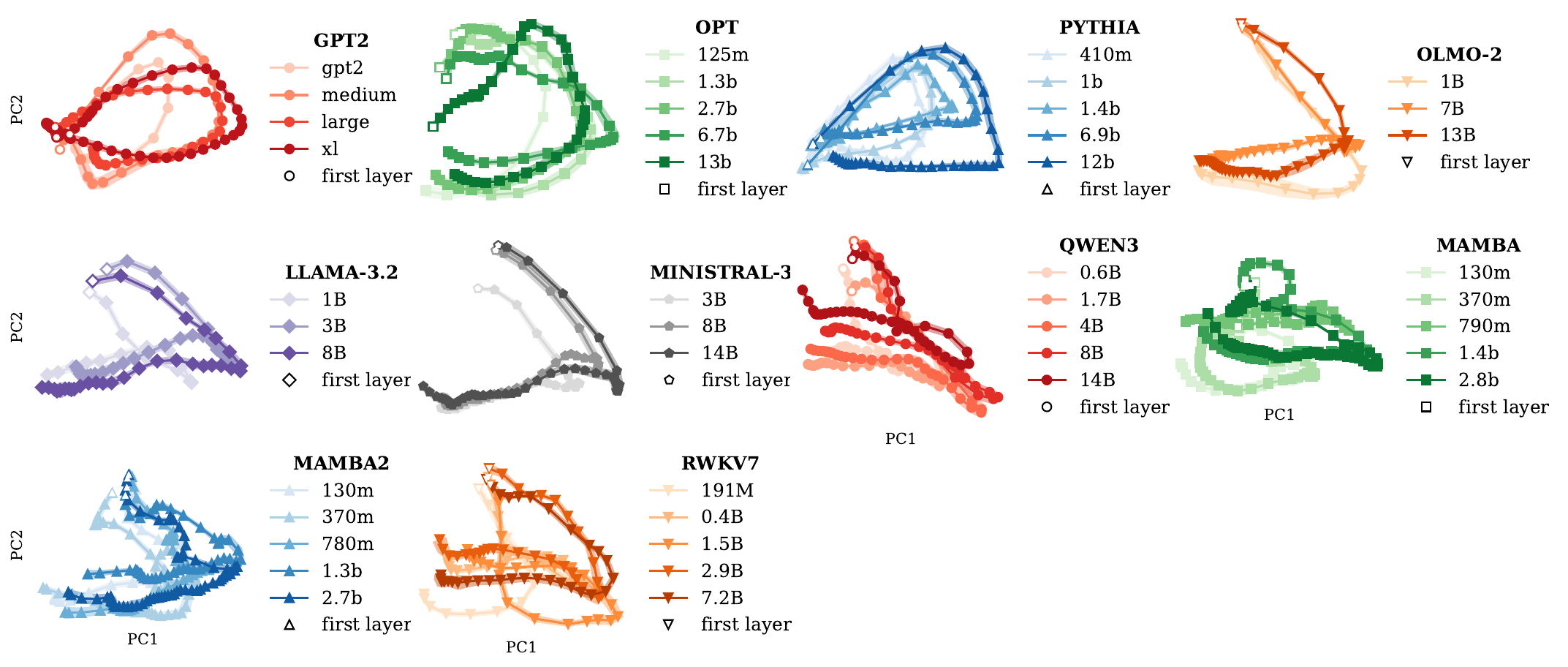}
    \caption{
        \textbf{Layer-wise Similarity for all LLMs:}
        To visualize how linguistic signatures change across layers, we project the layer-level signatures of all models into a shared PCA space\protect\footnotemark.
        Each marker represents one layer, successive layers from the same model are linked, and models are displayed family by family for readability.
        The main pattern is that models from the same family trace similar paths, indicating that they develop similar linguistic signatures across layers.
    }
    \label{fig:sim_matrices}
\end{figure*}

\footnotetext{
    For visualization purposes of this figure only, linguistic signatures were smoothed by applying a Gaussian filter with $\sigma=1$ along the layers of each model per feature before computing the PCA.
}

\section{Results}
\label{sec:results}

We trained an MLEM for each layer of all the 43 models.
\cref{fig:MLEM_perf} in Appendix shows the performance of these MLEMs measured with Spearman correlation on the held-out splits.
Across models, encoding performance follows the same broad profile: Spearman correlation is low in shallow and deep layers ($\approx 0.2$) and peaks in middle layers (0.5-0.7).
This is consistent with prior work finding that syntactic information is primarily processed in middle layers of large language models \citep{hewitt_structural_2019,diego-simon_polar_2024}.
Within each family, the peak tends to occur at a similar relative layer position across model sizes, but this position differs across families.

\subsection{Linguistic Signatures}

We first examined the linguistic signatures themselves and how they evolve across layers (\cref{fig:FIs} in Appendix).
Three main patterns emerge from this analysis.

Linguistic features show different importance dynamics across model layers.
This validates the sanity of the method and shows that features do not organize the sentence-level representations at the same layer depths.

In many models, the main syntactic features of the dataset, namely \textit{Relative Clause type} and \textit{Attachment site}, have their highest importance in middle layers.
This pattern mirrors the global encoding-performance profile and recovers again prior findings that syntactic information is primarily represented in middle layers of language models.

Models within the same family show visually similar linguistic signatures, even when they differ substantially in parameter count and number of layers.
This suggests that linguistic signatures are shaped by model family rather than scale.

\begin{figure*}[ht!]
    \centering
    \begin{subfigure}{.48\textwidth}
        \centering
        \includegraphics[width=\linewidth]{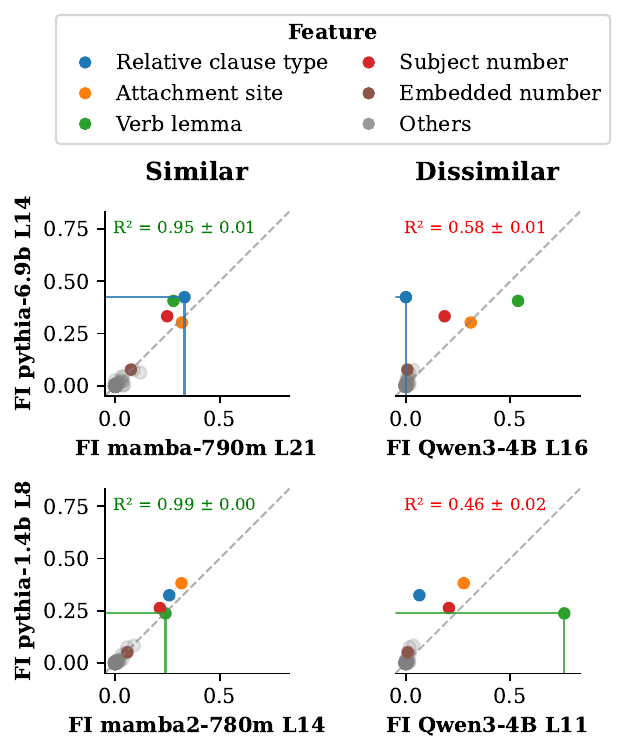}
        \caption{Feature Importances for reference layers against those of layers from other families with the closest and farthest linguistic signatures.}
    \end{subfigure}
    \hfill
    \begin{subfigure}{.48\textwidth}
        \centering
        \vfill
        \includegraphics[width=\linewidth]{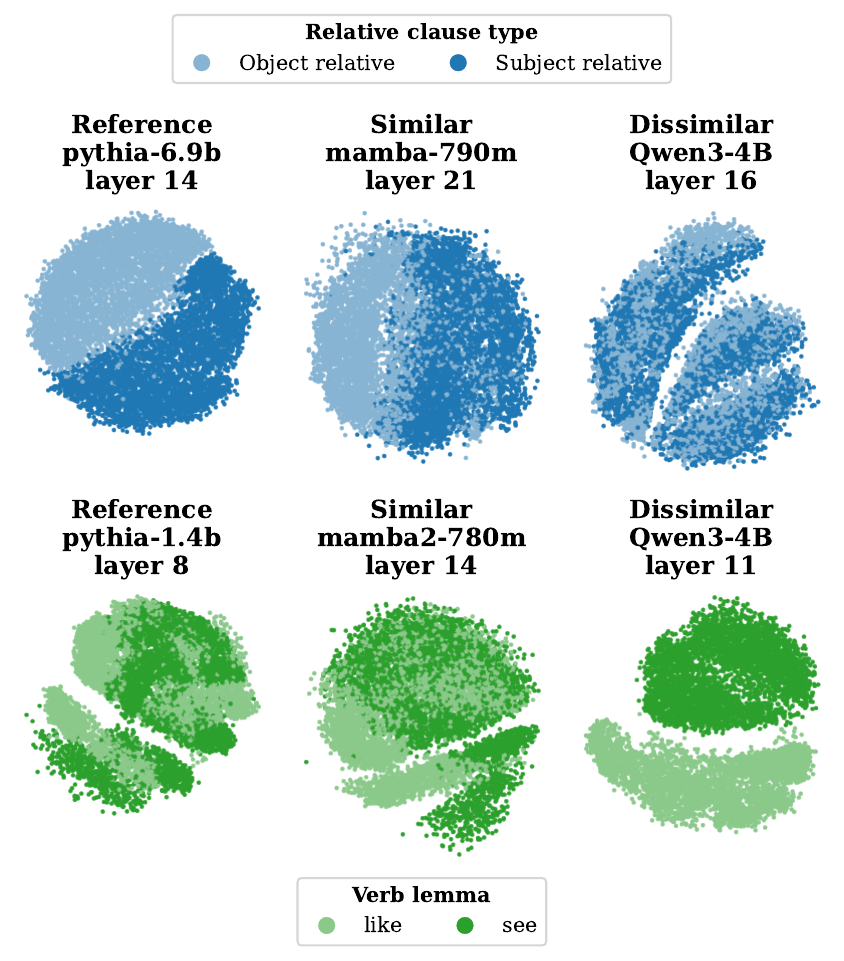}
        \vfill
        \caption{MDS projections of the neural representations of the reference, similar, and dissimilar layers, colored by the feature leading the similarity in each row.}
    \end{subfigure}
    \caption{
        \textbf{What Makes Two Linguistic Signatures Similar?}
        To keep the feature-level comparison tractable, we retained one layer per eighth of relative layer position for each model.
        For each retained layer, we compared retained layers from other model families in the same eighth and selected the layers with the closest and farthest linguistic signatures.
        The top row compares Pythia-6.9B layer 14 with Mamba-790M layer 21 and Qwen3-4B layer 16.
        \textit{Relative Clause type} dominates the reference and similar signatures and yields clear clustering in the corresponding neural spaces, whereas it has low importance and weak clustering in the dissimilar Qwen3-4B layer.
        The bottom row compares Pythia-1.4B layer 8 with Mamba2-780M layer 14 and Qwen3-4B layer 11.
        Here \textit{Verb lemma} contributes moderately in the reference and similar layers, but much more strongly in the dissimilar Qwen3-4B layer, producing correspondingly sharper clustering.
    }
    \vspace{-0.275cm}
    \label{fig:scatters}
\end{figure*}

\subsection{What Makes Two Models Think Alike?}

We asked which language models have the most similar linguistic signatures.
To do so, we compared their layer-wise linguistic-signature profiles using Dynamic Time Warping (DTW).
The resulting similarity structure is summarized with an MDS projection in \cref{fig:dtw_similarity}a and the full DTW distance matrix is shown in \cref{fig:dtw_heatmap} in Appendix.
We see a broad axis corresponding to release date and again models visually cluster primarily by family, indicating that models from the same family tend to develop similar linguistic signatures across layers.
To explain this structure quantitatively, we then treated the pairwise DTW matrix itself as the target of a second MLEM defined over model properties.
The right panel of \cref{fig:dtw_similarity} shows that release date, which acts as a proxy to summarize general LLM development, is the strongest predictor of model-level linguistic signature similarity, followed by family membership, with smaller contributions from parameter count, training token count, and depth.
By contrast, architecture class and depth-to-width ratio contribute little once the other model properties are included.
These results suggest that similarity in model-level linguistic signatures is not primarily shaped by scale or architecture class.
This pattern is consistent with the Platonic Representation Hypothesis in showing that different architectures can share representational profiles across layers.
However we find that convergence is partial and structured by model family and model generation rather than collapsing to a single endpoint when scale increases.

Do models from the same family follow similar layer-wise trajectories of linguistic signatures?
To address this question, we examined layer-level linguistic signatures (FI vectors) with a PCA computed over all layers from all models (\cref{fig:sim_matrices}).
We also show a Euclidean distance matrix between layer-level linguistic signatures in \cref{fig:feature_similarity_heatmap} in Appendix.
In this shared PCA space, models from the same family trace remarkably similar trajectories, indicating that their linguistic signatures evolve similarly across layers.
For many families, these trajectories form loops, with the last layers returning close to the first layers.
This loop-like organization is visible even in models without tied input and output embeddings, where weight tying means sharing the input token embedding matrix with the output softmax matrix \citep{press_using_2017}.
Because models must map their internal representations back to token space for next-token prediction, this return toward early-layer signatures may reflect constraints imposed by the output objective.

We also compare our approach to a feature-agnostic RSA baseline applied directly to the neural representations of the models from the GPT-2, OPT and Pythia families.
\cref{fig:rsa_heatmap,fig:rsa_similarity} in Appendix show the RSA similarity matrix and its MDS projection.
The latter shows that models from the same family tend to follow similar trajectories, much like the PCA of linguistic signatures.
However, it does not clearly show the loop-like organization, even for GPT-2, whose input and output embeddings are tied.
More importantly, RSA remains purely geometric: it does not identify which linguistic features drive similarity and it does not support the model-level, layer-level, and feature-level analyses enabled by linguistic signatures.

\subsection{What Makes Two Linguistic Signatures Similar?}

Why do two model layers represent linguistic information similarly or differently?
We addressed this question by zooming in on specific Feature Importance (FI) differences between layers.
For a given pair of layers, we asked which linguistic features account for their similarity or discrepancy, and whether the same features visibly organize their neural spaces.
To choose informative pairs, we turned to the layer-level linguistic-signature distance matrix (\cref{fig:feature_similarity_heatmap} in Appendix).
Inspecting all layer pairs was impractical so for all models we kept only one layer per eighth of relative layer position.
For each of them, we considered other retained layers from models in other families within the same eighth of relative layer position, then selected the ones with the closest and farthest linguistic signatures.
\cref{fig:scatters} shows two illustrative comparisons from this systematic search.

The first row of \cref{fig:scatters}a shows the FIs for Pythia-6.9B layer 14 against those of its closest counterpart, Mamba-790M layer 21, and its farthest counterpart, Qwen3-4B layer 16.
This comparison dissociates architecture and scale from linguistic signature, since Qwen3-4B is closer to Pythia-6.9B in architecture and parameter count but much farther in linguistic-signature space.
The most important feature for the similar layers is \textit{Relative Clause type} and we see clear clustering in their neural spaces whereas the dissimilar Qwen3-4B layer shows little such organization (\cref{fig:scatters}b top row).
In the second row, Pythia-1.4B layer 8 and Mamba2-780M layer 14 are again close, with \textit{Verb lemma} moderately important in both cases, while Qwen3-4B layer 11 assigns much higher importance to this feature and shows correspondingly sharper clustering in neural space.
This analysis shows that the feature-based approach provides a clear, interpretable explanation for why two layers are similar or different in their linguistic processing.

\section{Summary and Conclusions}
\label{sec:summary_conclusions}

Language models share the same computational task of next-token prediction.
However, architectural differences can drastically change the algorithms and representations they use to process text.
In this paper, we introduced a framework to test whether different models converge on similar linguistic representations or instead process text in different ways.
While traditional approaches treat similarity as a purely geometric property, we introduced a feature-based notion of representational similarity based on explicit linguistic signatures derived from Metric-Learning Encoding Models.

We tested this framework on decoder Transformers, State-Space Models, and Recurrent Neural Networks.
At the model level, profiles of linguistic signatures are broadly organized by release date, serving as a proxy for general LLM development while architecture class seems to not be a predictive feature.
Analyses at the model and layer levels comparing these profiles show that models within the same family tend to organize linguistic information similarly.
Finally, at the feature level, the method makes it possible to explain particular similarities and differences by identifying which linguistic features dominate which translates into organization differences of the corresponding neural spaces.

These findings refine the kind of convergence hypothesized by the Platonic Representation Hypothesis.
They support that different architecture classes, including Transformers, SSMs, and RNNs, can exhibit similar linguistic signatures.
However, they do not support a simple picture in which scale alone drives all models toward one undifferentiated representational endpoint.
Instead, linguistic convergence is structured by model family and model generation.

Together, these results illustrate the value of feature-based approaches.
By decomposing neural spaces into symbolic dimensions, we obtain an interpretable account of how language models represent and process language.
This comparison based on linguistic signatures can naturally generalize to domains like speech and vision, and to comparing artificial networks with biological brains.

\section{Limitations}
\label{sec:limitations}

While our approach explains model similarity using clear linguistic features, it has limitations.
First, we use a synthetic dataset where the linguistic features are largely independent, but it limits the analysis to a specific set of syntactic constructions.
Future work should test whether the same conclusions hold on natural, unconstrained text.

Second, linguistic signatures are defined relative to the features included in the symbolic description of the stimuli.
A model might rely on variables outside this feature set.
In that case, MLEMs would still explain which of the specified features account for pairwise neural distances, but the resulting signature would only capture part of the model's representational organization.

Third, our analyses operate at the level of whole layers.
This is appropriate for comparing broad representational geometries, but it does not identify the specific units, circuits, or algorithms that produce those geometries.
A feature could be perfectly computed by a localized mechanism and therefore have a weak layer-level importance.
MLEMs should therefore be viewed as complementary to mechanistic, causal, and unit-level analyses: they add a feature-level account of neural distances, but they do not replace circuit-level explanations.

Our results also do not directly test the cross-modal part of the Platonic Representation Hypothesis.
Our analysis is restricted to language models and to a syntactic dataset.
Future work could apply MLEMs to paired text, speech, vision, and brain data using shared interpretable features, testing whether cross-modal alignment is driven by the same explicit variables rather than by raw geometric overlap alone.

\begin{ack}
    We are grateful to Malo Renaudin, Maxence Pajot, and Pablo Diego-Simón for their valuable support and insightful feedback.

    Funding: This work was performed using HPC resources from GENCI-IDRIS (Grant 2026-AD011016055R1).

    Competing interests: The authors declare no competing interests.

\end{ack}

\clearpage

\printbibliography

\clearpage

\appendix

\begin{center}
    \LARGE{\textbf{Supplementary Material}}
\end{center}

\setcounter{figure}{0}
\setcounter{table}{0}
\renewcommand{\thefigure}{S\arabic{figure}}
\renewcommand{\thetable}{S\arabic{table}}

\section{Broader Impact}
\label{sec:broader_impact}

This work introduces a framework for understanding the linguistic representations of large language models.
By shifting from purely geometric comparisons to feature-based analyses, our approach contributes to the growing field of representational interpretability.
Improving our understanding of how different architectures process language is a critical step toward making these systems more transparent, predictable, and safe.
Furthermore, explicitly identifying the linguistic features that drive model representations can eventually help researchers uncover and mitigate implicit biases encoded in neural spaces.

\section{Model Inventory and Experimental Setup}
\label{sec:model_inventory}


We included the GPT-2 \citep{radford_language_2019}, OPT \citep{zhang_opt_2022}, and Pythia \citep{biderman_pythia_2023} families as older decoder-only Transformer baselines, because they represent different stages of open next-token prediction models, from early models trained on relatively modest corpora to systematic scaling suites.
We also included newer families trained with more recent methods and larger datasets, namely OLMo-2 \citep{olmo_2_2025}, Llama-3.2 \citep{grattafiori_llama_2024}, Ministral-3 \citep{liu_ministral_2026}, and Qwen3 \citep{yang_qwen3_2025}.
These newer families were chosen for complementary reasons: OLMo-2 for its transparent open training recipe \citep{olmo_2_2025}, Llama-3.2 and Ministral-3 as recent distilled families \citep{grattafiori_llama_2024, liu_ministral_2026}, and Qwen3 for very large-scale multilingual pretraining.
Finally, we included Mamba and Mamba-2 \citep{gu_mamba_2024a, dao_transformers_2024} and RWKV-7 \citep{peng_rwkv7_2025} to test whether similarities in linguistic structure persist across state-space and recurrent architectures.

We collected sentence representations as the hidden state from each layer on the last token (the period).
Due to causal attention in autoregressive Transformers and the recurrent nature of SSMs and RNNs, the last token is the only token that has seen the full sentence.
Therefore, only the last-token embedding can encode information from the whole sentence, including the linguistic features of interest.
We also tested representations as the concatenation of all token embeddings.
This richer representation was not retained because pairwise distances were driven by specific token identities rather than by linguistic features.


\begin{table}[!ht]
    \centering
    \renewcommand{\arraystretch}{1.05}
    \caption{
        \textbf{Evaluated language models.}
        Parameter counts are approximate and follow the checkpoint naming convention when available.
        Training token counts are taken either from the cited family paper or the corresponding Hugging Face model card.
        Note that we integrated Llama-3.1-8B to the Llama-3.2 family, as the latter doesn't have a text variant in this size range and the differences between those families are minimal.
    }
    \vspace{0.25cm}
    \label{tab:model_inventory}
    \begin{tabular*}{\linewidth}{@{\extracolsep{\fill}}c l c c c@{}}
        \toprule
        \textbf{Family} & \textbf{HuggingFace model} & \textbf{\#Params} & \textbf{Training tokens} & \textbf{Reference} \\
        \midrule
        \multirow{4}{*}{\makecell[c]{GPT-2\\\textcolor{lightgray}{Transformer}}} & \hfmodelshort{openai-community/gpt2}{gpt2} & 117M & \multirow{4}{*}{\makecell[c]{$\sim$10B\\(40GB WebText)}} & \multirow{4}{*}{\citealp{radford_language_2019}} \\
        & \hfmodelshort{openai-community/gpt2-medium}{gpt2-medium} & 345M & & \\
        & \hfmodelshort{openai-community/gpt2-large}{gpt2-large} & 762M & & \\
        & \hfmodelshort{openai-community/gpt2-xl}{gpt2-xl} & 1.5B & & \\
        \cmidrule(lr){1-5}
        \multirow{5}{*}{\makecell[c]{OPT\\\textcolor{lightgray}{Transformer}}} & \hfmodelshort{facebook/opt-125m}{opt-125m} & 125M & \multirow{5}{*}{180B} & \multirow{5}{*}{\citealp{zhang_opt_2022}} \\
        & \hfmodelshort{facebook/opt-1.3b}{opt-1.3b} & 1.3B & & \\
        & \hfmodelshort{facebook/opt-2.7b}{opt-2.7b} & 2.7B & & \\
        & \hfmodelshort{facebook/opt-6.7b}{opt-6.7b} & 6.7B & & \\
        & \hfmodelshort{facebook/opt-13b}{opt-13b} & 13B & & \\
        \cmidrule(lr){1-5}
        \multirow{5}{*}{\makecell[c]{Pythia\\\textcolor{lightgray}{Transformer}}} & \footnotesize{\hfmodelshort{EleutherAI/pythia-410m-deduped}{pythia-410m-deduped}} & 410M & \multirow{5}{*}{207B} & \multirow{5}{*}{\citealp{biderman_pythia_2023}} \\
        & \footnotesize{\hfmodelshort{EleutherAI/pythia-1b-deduped}{pythia-1b-deduped}} & 1B & & \\
        & \footnotesize{\hfmodelshort{EleutherAI/pythia-1.4b-deduped}{pythia-1.4b-deduped}} & 1.4B & & \\
        & \footnotesize{\hfmodelshort{EleutherAI/pythia-6.9b-deduped}{pythia-6.9b-deduped}} & 6.9B & & \\
        & \footnotesize{\hfmodelshort{EleutherAI/pythia-12b-deduped}{pythia-12b-deduped}} & 12B & & \\
        \cmidrule(lr){1-5}
        \multirow{3}{*}{\makecell[c]{OLMo-2\\\textcolor{lightgray}{Transformer}}} & \hfmodelshort{allenai/OLMo-2-0425-1B}{OLMo-2-0425-1B} & 1B & 4T & \multirow{3}{*}{\citealp{olmo_2_2025}} \\
        & \hfmodelshort{allenai/OLMo-2-1124-7B}{OLMo-2-1124-7B} & 7B & 4T & \\
        & \hfmodelshort{allenai/OLMo-2-1124-13B}{OLMo-2-1124-13B} & 13B & 5T & \\
        \cmidrule(lr){1-5}
        \multirow{3}{*}{\makecell[c]{Llama-3.2\\\textcolor{lightgray}{Transformer}}} & \hfmodelshort{meta-llama/Llama-3.2-1B}{Llama-3.2-1B} & 1B & $\leq$9T & \multirow{3}{*}{\citealp{grattafiori_llama_2024}} \\
        & \hfmodelshort{meta-llama/Llama-3.2-3B}{Llama-3.2-3B} & 3B & $\leq$9T & \\
        & \hfmodelshort{meta-llama/Llama-3.1-8B}{Llama-3.1-8B} & 8B & $\geq$15T & \\
        \cmidrule(lr){1-5}
        \multirow{3}{*}{\makecell[c]{Ministral-3\\\textcolor{lightgray}{Transformer}}} & \scriptsize{\hfmodelshort{mistralai/Ministral-3-3B-Base-2512}{Ministral-3-3B-Base-2512}} & 3B & \multirow{3}{*}{1-3T} & \multirow{3}{*}{\citealp{liu_ministral_2026}} \\
        & \scriptsize{\hfmodelshort{mistralai/Ministral-3-8B-Base-2512}{Ministral-3-8B-Base-2512}} & 8B & & \\
        & \scriptsize{\hfmodelshort{mistralai/Ministral-3-14B-Base-2512}{Ministral-3-14B-Base-2512}} & 14B & & \\
        \cmidrule(lr){1-5}
        \multirow{5}{*}{\makecell[c]{Qwen3\\\textcolor{lightgray}{Transformer}}} & \hfmodelshort{Qwen/Qwen3-0.6B-Base}{Qwen3-0.6B-Base} & 0.6B & \multirow{5}{*}{36T} & \multirow{5}{*}{\citealp{yang_qwen3_2025}} \\
        & \hfmodelshort{Qwen/Qwen3-1.7B-Base}{Qwen3-1.7B-Base} & 1.7B & & \\
        & \hfmodelshort{Qwen/Qwen3-4B-Base}{Qwen3-4B-Base} & 4B & & \\
        & \hfmodelshort{Qwen/Qwen3-8B-Base}{Qwen3-8B-Base} & 8B & & \\
        & \hfmodelshort{Qwen/Qwen3-14B-Base}{Qwen3-14B-Base} & 14B & & \\
        \cmidrule(lr){1-5}
        \multirow{5}{*}{\makecell[c]{Mamba\\\textcolor{lightgray}{SSM}}} & \hfmodelshort{state-spaces/mamba-130m-hf}{mamba-130m-hf} & 130M & \multirow{5}{*}{300B} & \multirow{5}{*}{\citealp{gu_mamba_2024a}} \\
        & \hfmodelshort{state-spaces/mamba-370m-hf}{mamba-370m-hf} & 370M & & \\
        & \hfmodelshort{state-spaces/mamba-790m-hf}{mamba-790m-hf} & 790M & & \\
        & \hfmodelshort{state-spaces/mamba-1.4b-hf}{mamba-1.4b-hf} & 1.4B & & \\
        & \hfmodelshort{state-spaces/mamba-2.8b-hf}{mamba-2.8b-hf} & 2.8B & & \\
        \cmidrule(lr){1-5}
        \multirow{5}{*}{\makecell[c]{Mamba-2\\\textcolor{lightgray}{SSM}}} & \hfmodelshort{AntonV/mamba2-130m-hf}{mamba2-130m-hf} & 130M & \multirow{5}{*}{300B} & \multirow{5}{*}{\citealp{dao_transformers_2024}} \\
        & \hfmodelshort{AntonV/mamba2-370m-hf}{mamba2-370m-hf} & 370M & & \\
        & \hfmodelshort{AntonV/mamba2-780m-hf}{mamba2-780m-hf} & 780M & & \\
        & \hfmodelshort{AntonV/mamba2-1.3b-hf}{mamba2-1.3b-hf} & 1.3B & & \\
        & \hfmodelshort{AntonV/mamba2-2.7b-hf}{mamba2-2.7b-hf} & 2.7B & & \\
        \cmidrule(lr){1-5}
        \multirow{5}{*}{\makecell[c]{RWKV-7\\\textcolor{lightgray}{RNN}}} & \hfmodelshort{fla-hub/rwkv7-191M-world}{rwkv7-191M-world} & 191M & 1.6T & \multirow{5}{*}{\citealp{peng_rwkv7_2025}} \\
        & \hfmodelshort{fla-hub/rwkv7-0.4B-world}{rwkv7-0.4B-world} & 0.4B & 3.1T & \\
        & \hfmodelshort{fla-hub/rwkv7-1.5B-world}{rwkv7-1.5B-world} & 1.5B & 5.6T & \\
        & \hfmodelshort{fla-hub/rwkv7-2.9B-world}{rwkv7-2.9B-world} & 2.9B & 5.6T & \\
        & \hfmodelshort{fla-hub/rwkv7-7.2B-g0a}{rwkv7-7.2B-g0a} & 7.2B & 5.6T & \\
        \bottomrule
    \end{tabular*}
\end{table}

\textbf{OPT:} We excluded \hfmodel{facebook/opt-350m} because its architecture uses heterogeneous hidden state dimensions (1024 to 512 in the final layer), which prevents the uniform layer-wise analysis applied to all other models.

\textbf{Pythia:} We excluded \hfmodel{EleutherAI/pythia-14m-deduped}, \hfmodel{EleutherAI/pythia-70m-deduped}, and \hfmodel{EleutherAI/pythia-160m-deduped} because exploratory MLEM fits stayed near floor across layers and did not reliably recover the \textit{Relative Clause type} and \textit{Attachment site} contrasts.
These models therefore appeared too small to produce stable sentence-level syntactic structure for the present analysis.
We also excluded \hfmodel{EleutherAI/pythia-2.8b-deduped} because its FI profiles were degenerate and inconsistent with neighboring Pythia checkpoints, so we suspect a corrupted checkpoint or failed training.

\FloatBarrier
\section{Relative Clause Dataset}
\label{sec:relative_clause_dataset}
\begin{figure}[h!]
    \centering
    \includegraphics[width=\linewidth]{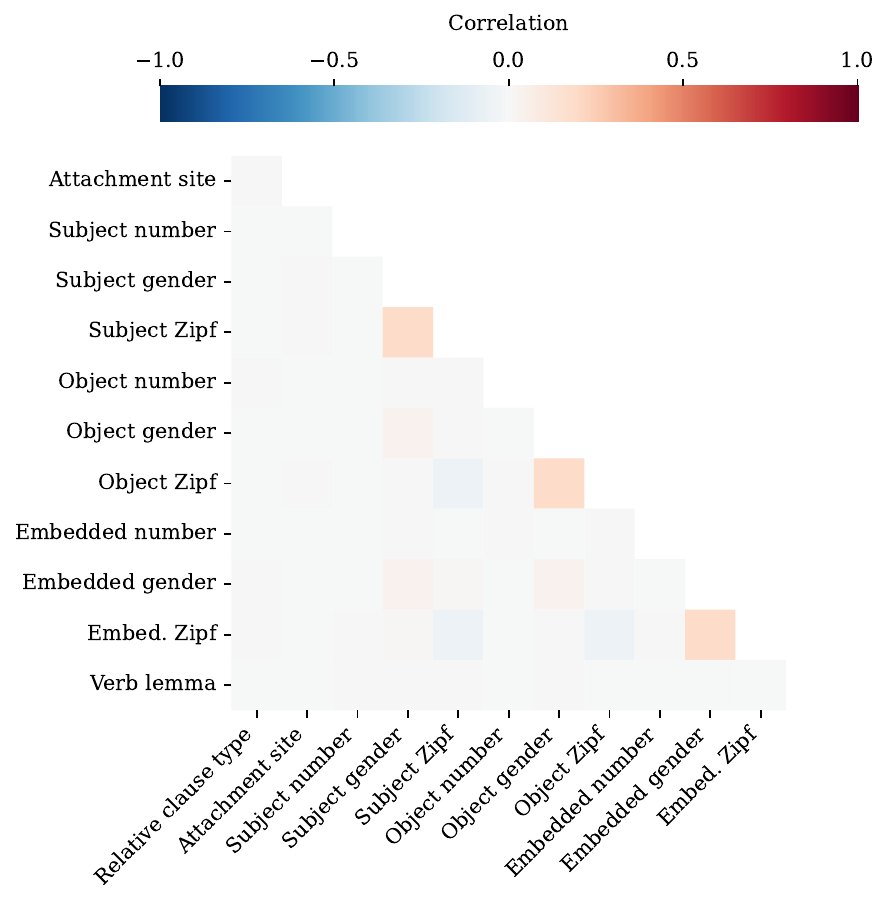}
    \caption{
        \textbf{Linguistic features in the Relative Clause dataset are largely uncorrelated.}
        This heatmap shows the Pearson correlations between the pairwise distance matrices of the 12 linguistic features in the dataset.
        The correlations are very low so the dataset is well-balanced and the Feature Importances are reliable.
    }
    \label{fig:rc_correlations}
\end{figure}

This section details the synthetic Relative Clause dataset used throughout the study.
We detail its 2x2 design in \cref{tab:rc_design}, its linguistic features in \cref{tab:rc_features}, and the correlations between them in \cref{fig:rc_correlations}.
This dataset is a derived version of the Relative Clause dataset introduced by \citet{jalouzot_metric_2025}.
Relative to the original release, we removed the intervener-number and intervener-gender columns, used a simplified noun lexicon and the verbs \textit{see} and \textit{like} to instantiate the templates, and recomputed Zipf-frequency values from those lexical items.

\begin{table}[!ht]
    \centering
    \renewcommand{\arraystretch}{1.5}
    \caption{\textbf{Relative Clause dataset design.} 2x2 design emphasizing center-embedded vs peripheral and subject vs object relative clauses (adapted from \citealp{jalouzot_metric_2025}).}
    \vspace{0.25cm}
    \begin{tabular}{ccl}
        \toprule
        \multirow{1}{*}{\textbf{Attachment Site}}
        & \multirow{1}{*}{\textbf{Relative Clause Type}}
        & \multirow{1}{*}{\textbf{Template / Example}} \\
        \toprule
        \multirow{4}{*}{\textbf{Center-Embedded}}
        & \multirow{2}{*}{\textbf{Subject relative}}
        & \textit{Subj \textcolor{purple}{[who verb obj]} verb obj.} \\
        & & \textcolor{gray}{The woman \textcolor{purple}{[who sees the girl]} likes the queen.} \\
        \cline{2-3}
        & \multirow{2}{*}{\textbf{Object relative}}
        & \textit{Subj \textcolor{purple}{[who subj verb]} verb obj.} \\
        & & \textcolor{gray}{The woman \textcolor{purple}{[who the girl sees]} likes the queen.} \\
        \midrule
        \multirow{4}{*}{\makecell[t]{\textbf{Peripheral} \\ \textbf{(Right-Branching)}}}
        & \multirow{2}{*}{\textbf{Subject relative}}
        & \textit{Subj verb obj \textcolor{purple}{[who verb obj]}.} \\
        & & \textcolor{gray}{The woman sees the girl \textcolor{purple}{[who likes the queen]}.} \\
        \cline{2-3}
        & \multirow{2}{*}{\textbf{Object relative}}
        & \textit{Subj verb obj \textcolor{purple}{[who subj verb]}.} \\
        & & \textcolor{gray}{The woman likes the girl \textcolor{purple}{[who the queen sees]}.} \\
        \bottomrule
    \end{tabular}
    \label{tab:rc_design}
\end{table}

\begin{table}[h!]
    \centering
    \renewcommand{\arraystretch}{1.45}
    \caption{\textbf{Linguistic features in the Relative Clause dataset} (adapted from \citealp{jalouzot_metric_2025}).}
    \vspace{0.25cm}
    \begin{tabularx}{\linewidth}{@{}ll>{\raggedright\arraybackslash}X@{}}
        \toprule
        \textbf{Feature} & \textbf{Values} & \textbf{Description} \\
        \midrule
        \textbf{Relative Clause type} & Subject relative, Object relative & RC modifies subject or object of main clause.\\\midrule
        \textbf{Attachment site} & Peripheral, Center-embedded & Position of RC relative to main clause.\\\midrule
        \textbf{Subject number} & Singular, Plural & Grammatical number of main clause subject.\\\midrule
        \textbf{Subject gender} & Feminine, Masculine & Grammatical gender of main clause subject.\\\midrule
        \textbf{Subject frequency} & 4.86, 5.17, 5.35, 5.38, 5.82 & Zipf frequency of main clause subject.\\\midrule
        \textbf{Object number} & Singular, Plural & Grammatical number of main clause object.\\\midrule
        \textbf{Object gender} & Feminine, Masculine & Grammatical gender of main clause object.\\\midrule
        \textbf{Object frequency} & 4.86, 5.17, 5.35, 5.38, 5.82 & Zipf frequency of main clause object.\\\midrule
        \textbf{Embedded number} & Singular, Plural & Grammatical number of embedded clause subject or object.\\\midrule
        \textbf{Embedded gender} & Feminine, Masculine & Grammatical gender of embedded clause subject or object.\\\midrule
        \textbf{Embedded frequency} & 4.86, 5.17, 5.35, 5.38, 5.82 & Zipf frequency of embedded clause subject or object.\\\midrule
        \textbf{Verb lemma} & see, like & Lemma of the verb.\\
        \bottomrule
    \end{tabularx}
    \label{tab:rc_features}
\end{table}

\clearpage

\section{Model Properties Analysis}

\begin{figure}[h!]
    \centering
    \includegraphics[width=.8\linewidth]{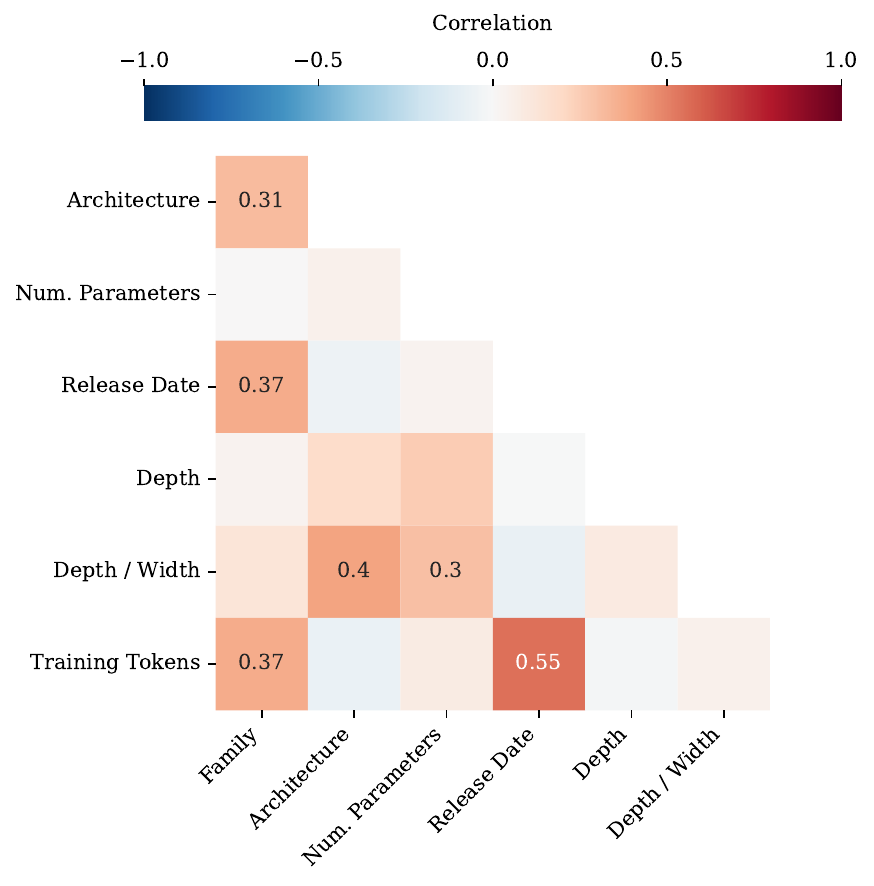}
    \caption{
        \textbf{Model properties are moderately correlated.}
        This heatmap shows the Pearson correlations between the pairwise distance matrices of the model properties used in the analysis of DTW similarities (\cref{fig:dtw_fi}).
        Width and vocabulary size were excluded from this analysis because they correlate too much with family membership and number of parameters which would compromise the interpretability of the Feature Importances.
    }
    \label{fig:dtw_feature_corrs}
\end{figure}

\clearpage

\section{Additional Results}
The remaining figures provide full visual outputs for layer-wise signatures, model-level similarity, baseline geometric similarity, and MLEM fit quality.

\begin{figure*}[ht!]
    \centering
    \includegraphics[width=\linewidth]{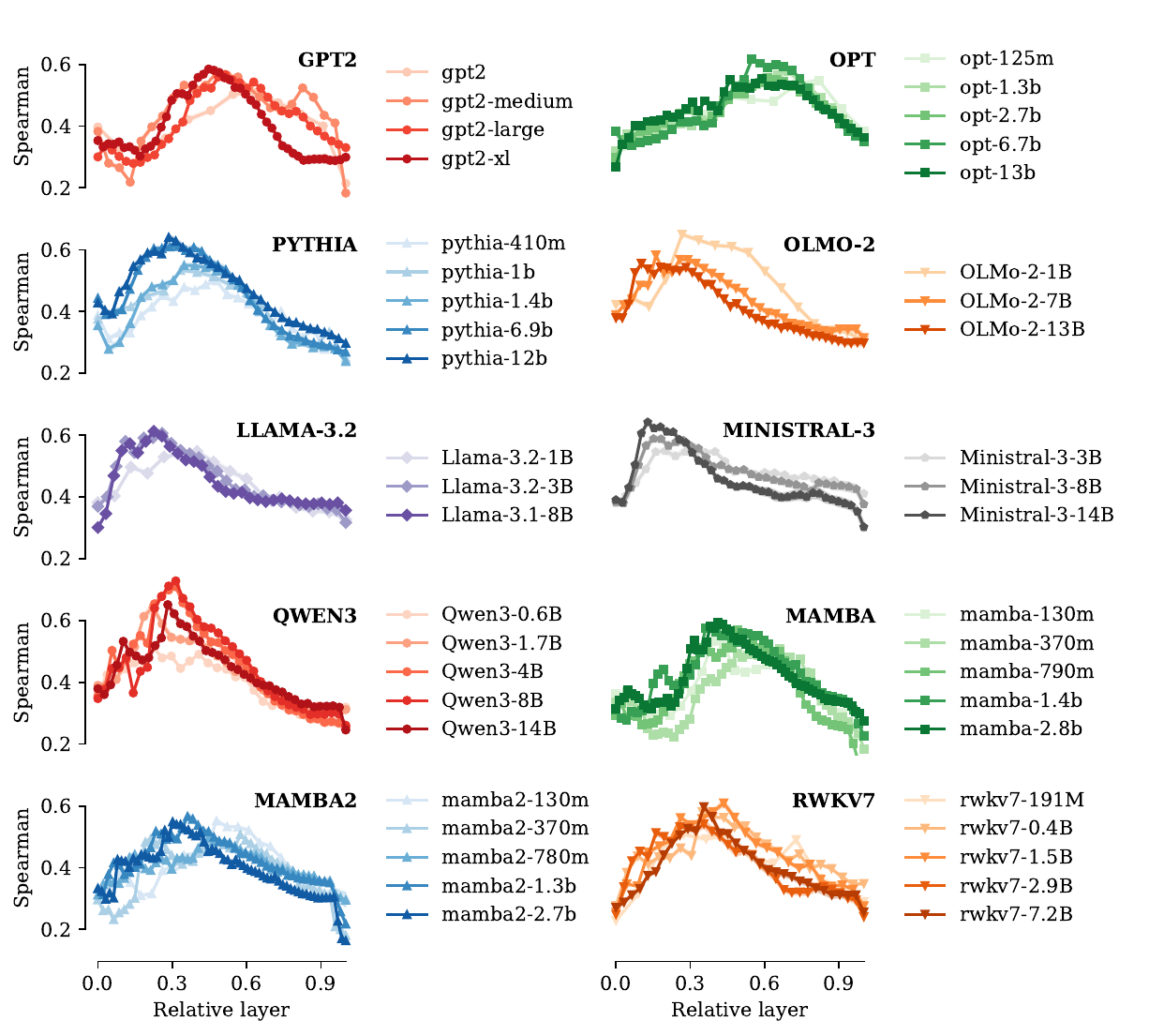}
    \caption{
        \textbf{Layer-wise MLEM performance.}
        Spearman correlation between predicted and actual neural distances for all evaluated models, shown as a function of relative layer position.
    }
    \label{fig:MLEM_perf}
\end{figure*}

\begin{figure*}[ht!]
    \thisfloatpagestyle{empty}
    \centering
    \vspace{-1cm}
    \includegraphics[height=\textheight]{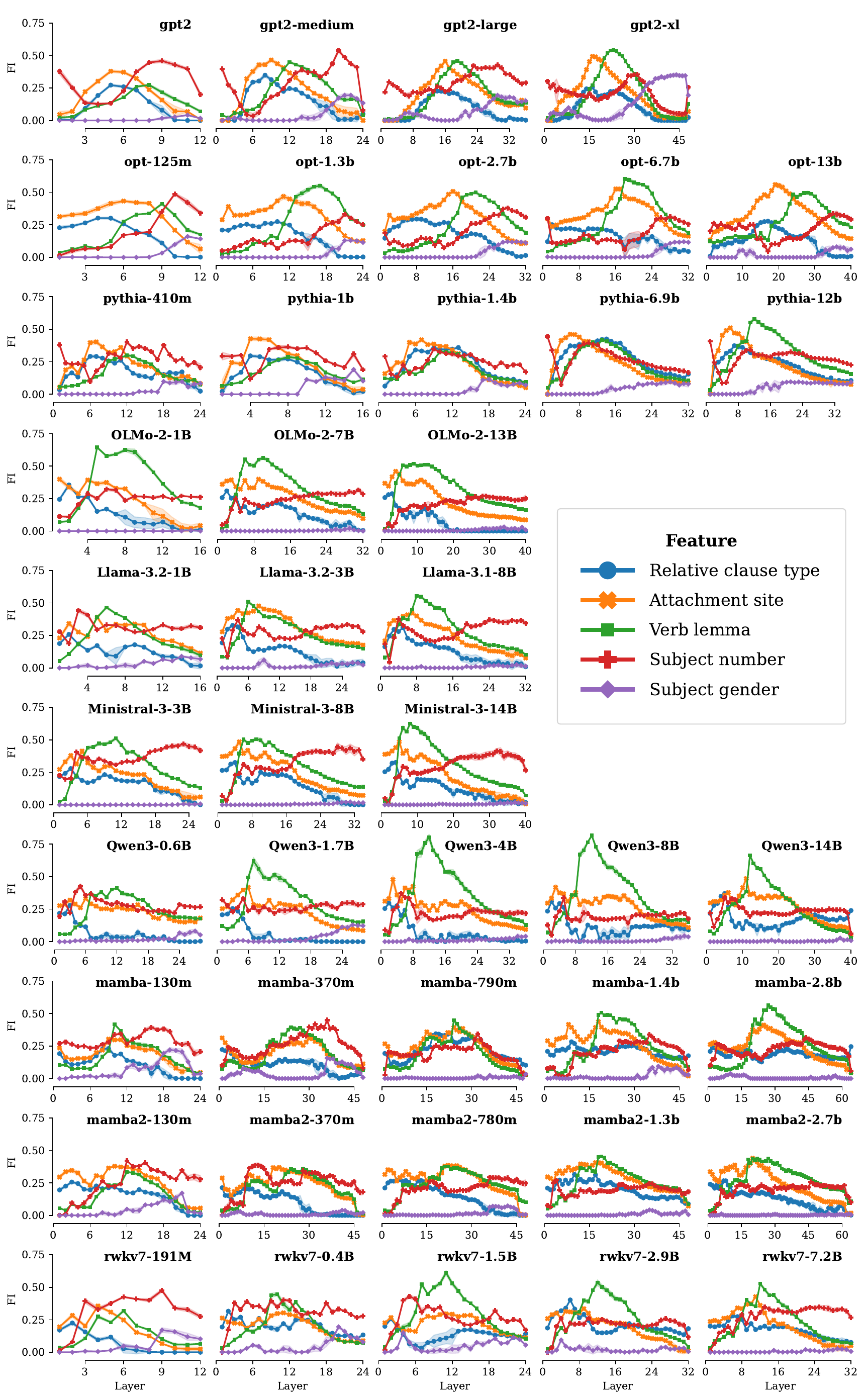}
    \caption{
        \textbf{Linguistic Signatures.}
        Metric-Learning Encoding Models measure the importance of each linguistic feature in explaining differences between sentence representations at each layer.
        Dominant features vary across layers, showing which linguistic rules structure the layer's neural space.
    }
    \label{fig:FIs}
\end{figure*}

\begin{figure*}[ht!]
    \centering
    \includegraphics[width=\linewidth]{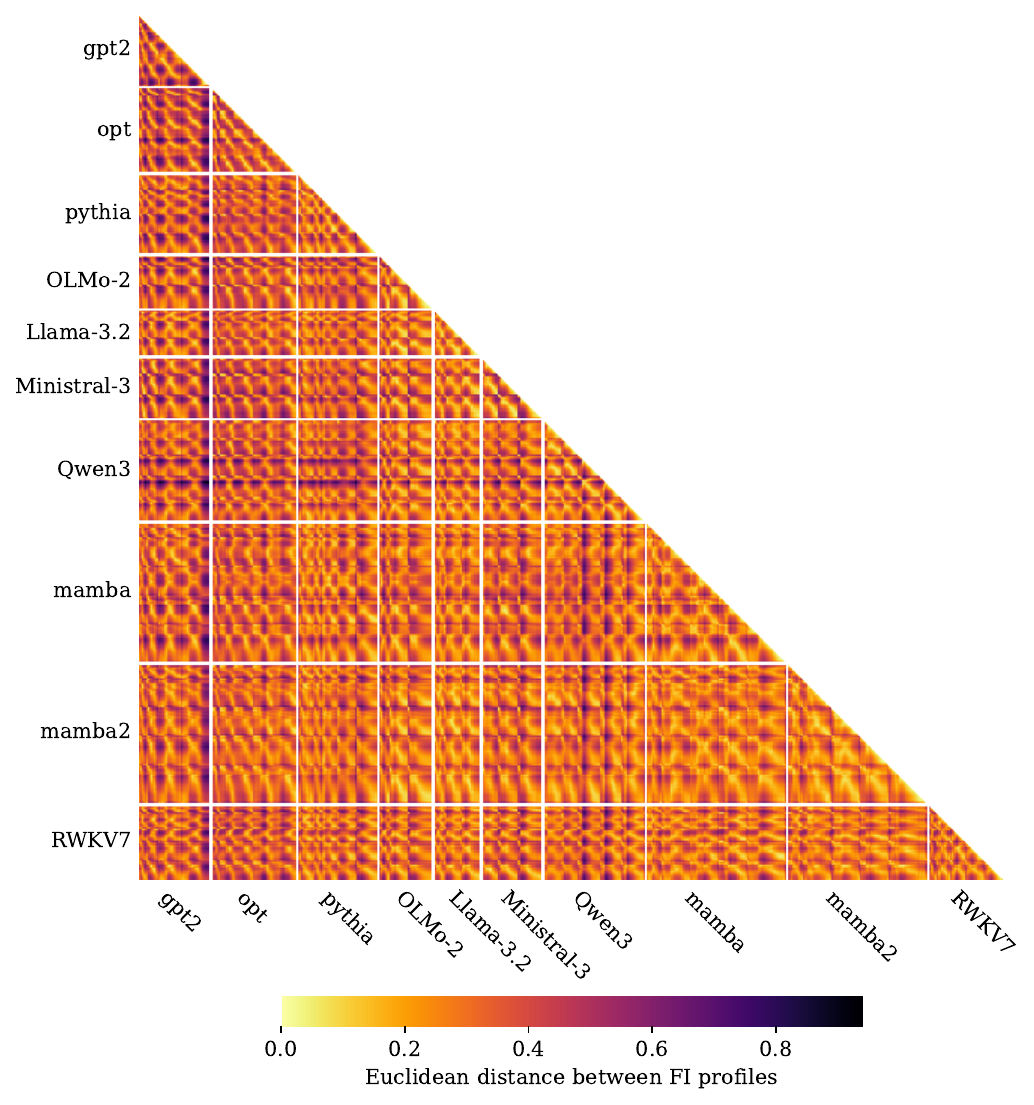}
    \caption{
        \textbf{Heatmap of layer-level linguistic signature similarity.}
        Each row and column corresponds to one model layer, ordered by model family, model size, and layer.
        Individual model and layer labels are omitted for readability and white gaps separate families.
        Colors show Euclidean distances between layer-level linguistic signatures, with smaller values indicating more similar FI profiles.
    }
    \label{fig:feature_similarity_heatmap}
\end{figure*}

\clearpage

To compute DTW distances, we use the implementation available in the Python package \verb+dtw-python+ (cf.~corresponding \href{https://github.com/DynamicTimeWarping/dtw-python}{GitHub repository}; \cite{giorgino_computing_2009}).
We used the returned \verb+normalizedDistance+ attribute.

\begin{figure*}[ht!]
    \centering
    \includegraphics[width=\linewidth]{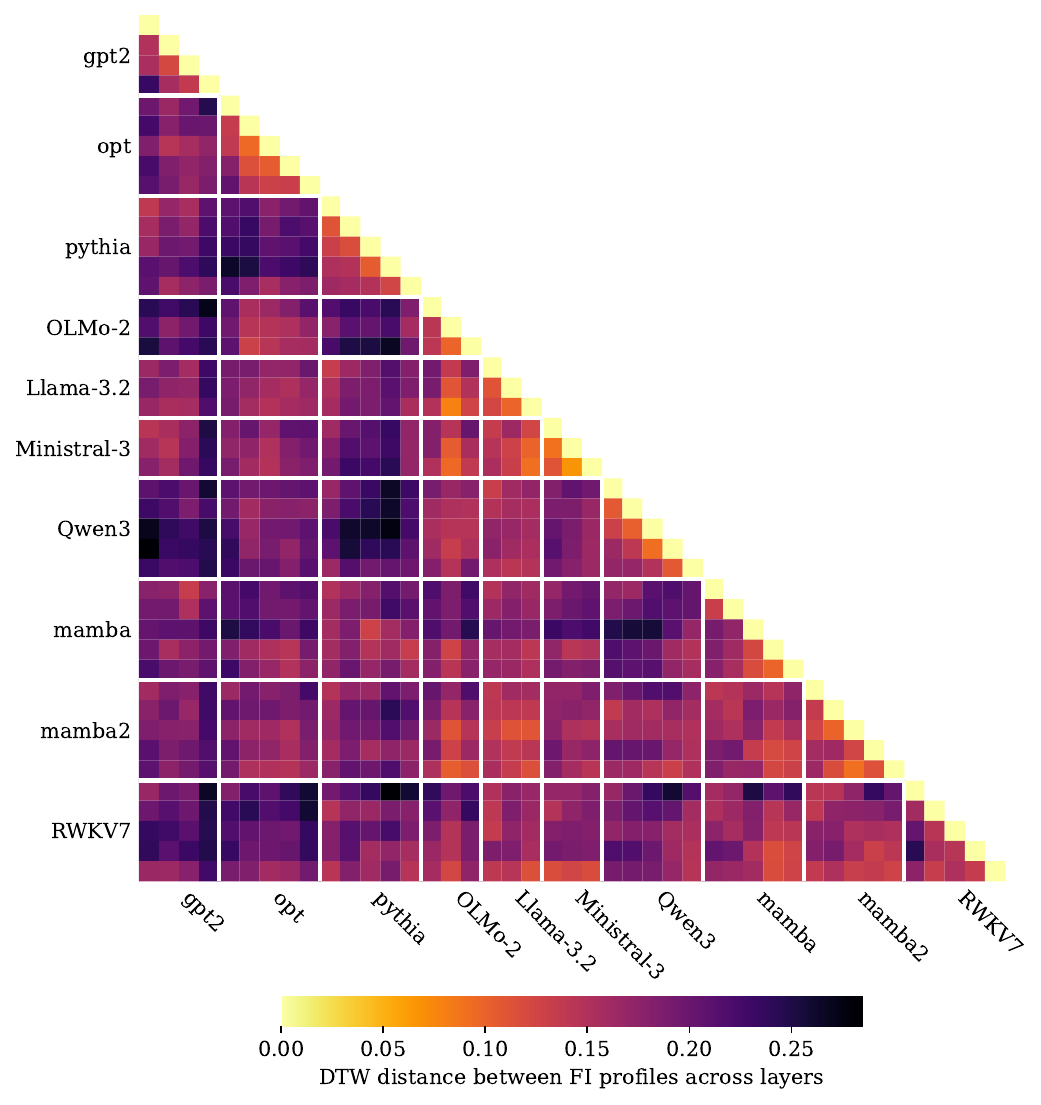}
    \caption{
        \textbf{DTW heatmap of model-level linguistic signature similarity.}
        Each row and column corresponds to one full model, ordered by family and model size.
        Individual model labels are omitted for readability and white gaps separate families.
        Colors show normalized Dynamic Time Warping distances between layer-wise FI profiles, with smaller values indicating more similar linguistic-signature trajectories.
    }
    \label{fig:dtw_heatmap}
\end{figure*}

\begin{figure*}[ht!]
    \centering
    \includegraphics[width=.8\linewidth]{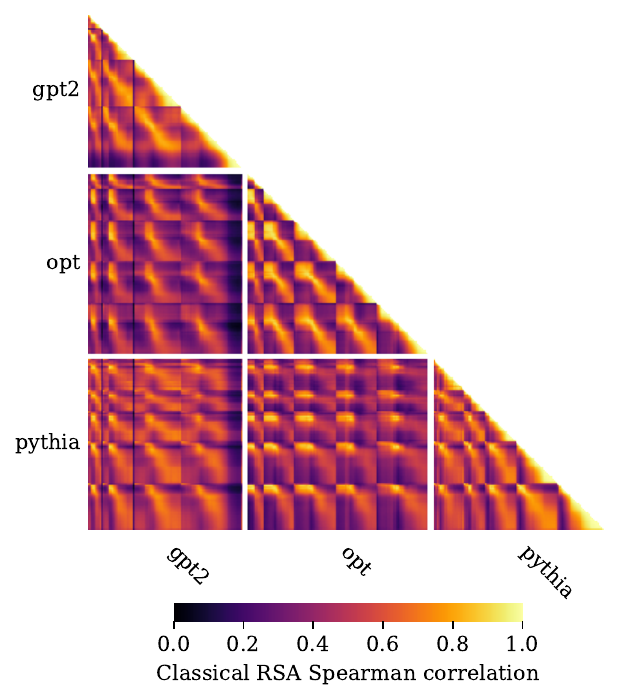}
    \caption{
        \textbf{RSA heatmap of raw geometric similarity.}
        Each row and column corresponds to one model layer from the GPT-2, OPT, and Pythia models included in this baseline, ordered by model family, model size, and layer.
        Individual layer labels are omitted for readability and white gaps separate families.
        Colors show Spearman correlations between pairwise neural-distance matrices computed directly from sentence-level representations, with larger values indicating more similar raw representational geometry.
    }
    \label{fig:rsa_heatmap}
\end{figure*}

\begin{figure*}[ht!]
    \centering
    \includegraphics[width=\linewidth]{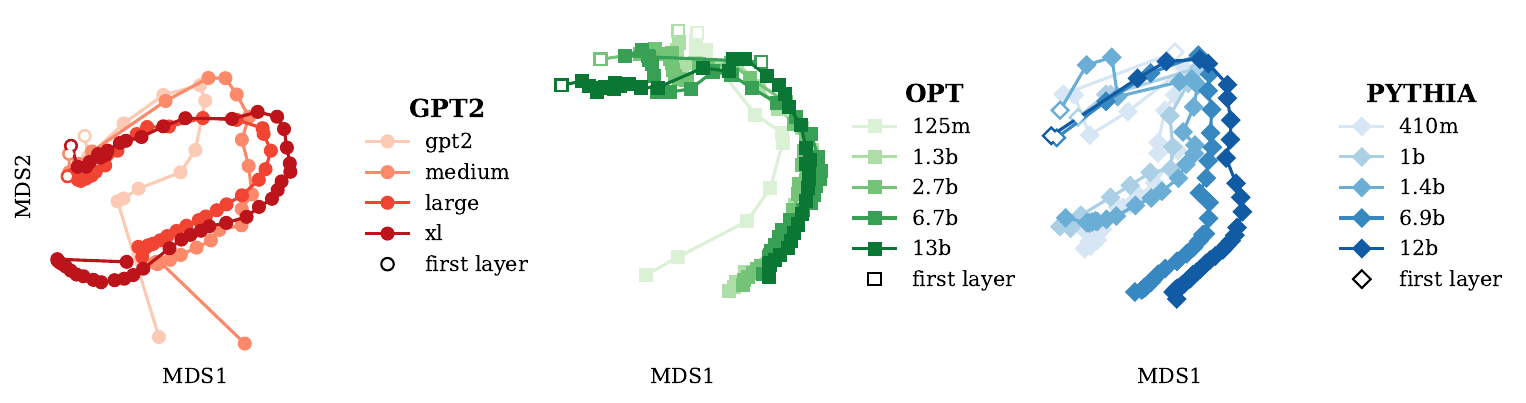}
    \caption{
        \textbf{Feature-agnostic RSA projection.}
        MDS projection of the RSA-based raw geometric similarity shown in \cref{fig:rsa_heatmap}, computed jointly over all layers from all models and displayed family by family for readability.
        Models from the same family still follow similar trajectories, but the loop-like organization is weaker than in the PCA of linguistic signatures.
    }
    \label{fig:rsa_similarity}
\end{figure*}

\clearpage

\section{Compute Resources}
\label{sec:compute_resources}
The project, including exploration and final experiments, required roughly 2,500 CPU hours and 500 H100 GPU hours on an HPC cluster.
The extraction of LLM embeddings was performed on H100 GPUs and the MLEM pipeline can leverage GPUs and was run on a mix of GPU and CPU workers.

\section{Asset Licenses}
\label{sec:asset_licenses}
The Relative Clause dataset used in this paper is a slight modification of the dataset introduced by \citet{jalouzot_metric_2025}.
The original dataset does not currently have a public license.
We obtained authorization from the original authors to modify and redistribute this version in the supplementary material code.
The public models used in the paper are released under the following terms: GPT-2 under MIT; OPT under the Meta OPT non-commercial research license; Pythia, OLMo-2, Ministral-3, Qwen3, Mamba, and RWKV-7 under Apache-2.0; Mamba-2 converted checkpoints under MIT; and Llama 3.1/3.2 under the Llama Community License and Acceptable Use Policy.
The main software dependencies are under permissive open-source licenses, except \texttt{dtw-python}, an external GPLv3 package used to compute the Dynamic Time Warping distance.
We use it as an external dependency without modifying its source code.



\end{document}